\documentclass[10pt,twocolumn,letterpaper]{article}

\usepackage{cvpr}              

\usepackage{amsmath}
\usepackage{kotex}

\usepackage{colortbl}  

\usepackage[accsupp]{axessibility}

\usepackage{multirow}

\definecolor{cvprblue}{rgb}{0.21,0.49,0.74}
\usepackage[pagebackref,breaklinks,colorlinks,allcolors=cvprblue]{hyperref}

\def\paperID{****} 
\def\confName{CVPR}
\def\confYear{2026}

\title{HypeVPR: Exploring Hyperbolic Space for Perspective to Equirectangular Visual Place Recognition}

\author{
Suhan Woo$^{1}$ \quad
Seongwon Lee$^{2}$ \quad
Jinwoo Jang$^{1}$ \quad
Euntai Kim$^{*1,3}$ \\[0.5em]
$^{1}$Yonsei University \quad
$^{2}$Kookmin University \quad
$^{3}$Korea Institute of Science and Technology (KIST)
}

\begin{document}


\twocolumn[{%
\maketitle
  \begin{center}
    \includegraphics[width=.98\linewidth]{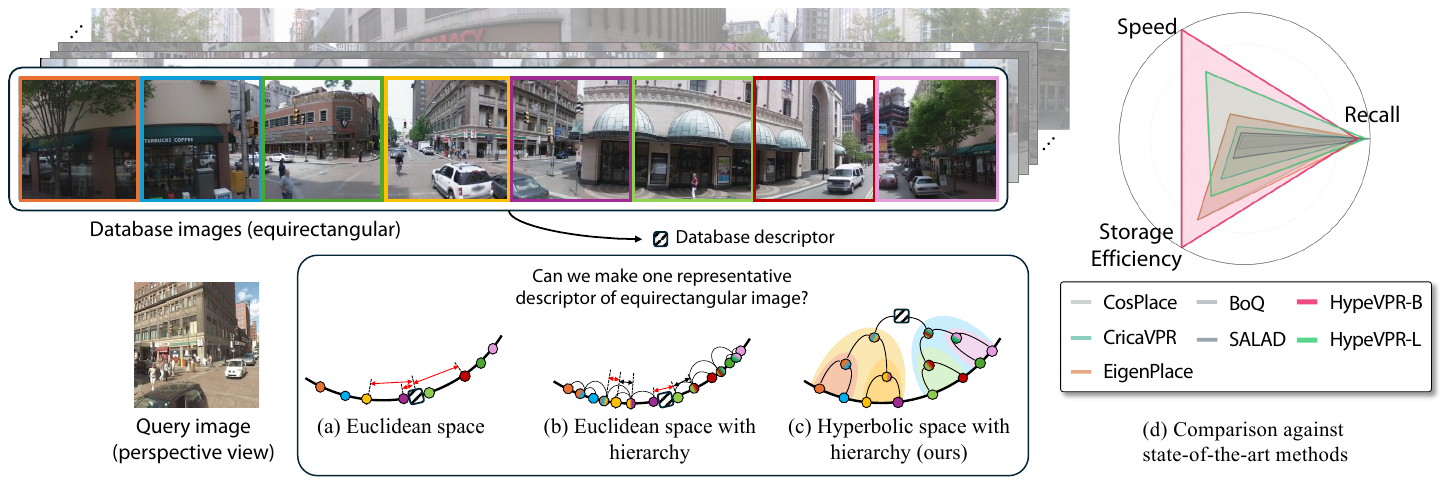}
    \captionof{figure}{The perspective to equirectangular visual place recognition (P2E VPR) problem requires matching a distinct query image with a panoramic database image.
The yellow-highlighted region (4th from the left) corresponds to the matching area, while the rest is dominated by redundant FoVs, making it difficult to generate a single representative descriptor.
(a) In Euclidean space, modeling the global descriptor that faithfully preserves relational distances among multiple fields of view (FoVs) is challenging.
(b) Even with hierarchical modeling, Euclidean representations suffer from geometric distortion that hinders accurate hierarchy preservation.
(c) Our hierarchical modeling in hyperbolic space naturally embeds multi-level relationships with minimal distortion, producing a compact representative descriptor that effectively integrates all FoVs within the equirectangular image.
(d) Comparison of speed, recall, and storage efficiency against state-of-the-art methods, showing the favorable trade-offs of our approach. (See Tab.~\ref{tab:sf_xl} for details.)}
    \label{fig:figure1}
  \end{center}
  }]

\begin{abstract}


Visual environments are inherently hierarchical, as a panoramic view naturally encompasses and organizes multiple perspective views within its field. Capturing this hierarchy is crucial for effective perspective-to-equirectangular (P2E) visual place recognition. In this work, we introduce HypeVPR, a hierarchical embedding framework in hyperbolic space specifically designed to address the challenges of P2E matching. HypeVPR leverages the intrinsic ability of hyperbolic space to represent hierarchical structures, allowing panoramic descriptors to encode both broad contextual information and fine-grained local details. To this end, we propose a hierarchical feature aggregation mechanism that organizes local-to-global feature representations within hyperbolic space. Furthermore, HypeVPR’s hierarchical organization naturally enables flexible control over the accuracy–efficiency trade-off without additional training, while maintaining robust matching across different image types. 
This approach enables HypeVPR to achieve competitive performance while significantly accelerating retrieval and reducing database storage requirements.
Project page: \hyperlink{https://suhan-woo.github.io/HypeVPR/}{https://suhan-woo.github.io/HypeVPR/}

\end{abstract}    
\section{Introduction}
\label{sec:intro}

Visual Place Recognition (VPR) aims to identify specific locations by retrieving the most visually similar images from a database given a query image, building upon image matching techniques in computer vision~\cite{arandjelovic2016netvlad, berton2022deep}. Conventional VPR methods were primarily based on perspective-to-perspective (P2P) matching and have demonstrated strong performance~\cite{berton2022rethinking,berton2023eigenplaces,ali2024boq,lu2024towards,lu2024cricavpr}. With these advances, VPR has gained increasing attention in mobile platforms such as autonomous robots and vehicles, as it serves as a key component for tasks like relocalization and kidnapping recovery~\cite{middelberg2014scalable,zhou2023lcpr,hane20173d,doan2019scalable}. However, its applicability to large real-world environments—especially for mobile systems—remains limited. This is primarily because the database must be densely sampled with view-specific images to cover all possible viewing directions for queries captured from arbitrary viewpoints. This requirement results in substantial storage demands and high retrieval costs.

A promising alternative is the perspective-to-equirectangular (P2E) framework, where the query is a perspective image while the database consists of panoramic equirectangular images. Panoramic representations can significantly reduce redundancy since each location can be represented by a single panorama rather than multiple directional views. However, achieving effective matching between perspective queries and panoramic database images remains highly challenging. Existing methods~\cite{orhan2021efficient,shi2023panovpr} generally follow a P2P pipeline and perform an exhaustive sliding-window search over the panorama, resulting in significant computational overhead without offering meaningful improvement over conventional P2P VPR. These limitations motivate the need for a more efficient approach that fully exploits panoramic structure.

Our motivation stems from the observation that visual environments exhibit inherent hierarchical structure~\cite{parikh2007hierarchical,choi2010exploiting,ge2023hyperbolic}. Panoramic views naturally encompass multiple perspective observations within a single scene, and these relationships can be effectively organized in a hierarchical structure. This insight motivates our approach to P2E VPR, which leverages such hierarchy to capture both broad global context and fine-grained local details in a structured, multi-level representation.

In this paper, we introduce HypeVPR, a novel hyperbolic embedding framework for P2E VPR. As illustrated in Fig.~\ref{fig:figure1}, hyperbolic space naturally models hierarchical relationships~\cite{sarkar2011low,sala2018representation,ge2023hyperbolic}, enabling compact encoding of broad contextual structure with minimal distortion which is difficult to achieve in Euclidean space. Building on this property, HypeVPR divides each panoramic view into regions with varying fields of view (FoVs) and organizes their features hierarchically in the embedding space: higher levels capture coarse global context, while lower levels encode fine, localized details. Based on this hierarchy, HypeVPR performs adaptive retrieval by selectively activating descriptors at different levels, enabling a flexible balance between accuracy and efficiency. As a result, HypeVPR achieves both robust and efficient P2E VPR, while offering controllable trade-offs between retrieval accuracy and computational cost, as illustrated in Fig.~\ref{fig:figure1}(d).

Our extensive evaluation shows that HypeVPR delivers comparable recognition accuracy to state-of-the-art models while maintaining superior efficiency in terms of retrieval speed and storage.
Our main contributions are summarized as follows:
\begin{itemize}[leftmargin=5mm]
\setlength{\itemsep}{1pt}
\setlength{\parskip}{0pt}
\setlength{\parsep}{0pt}
\item We present a hyperbolic space-based VPR framework specifically designed for perspective-to-equirectangular (P2E) matching.
\item We propose a hierarchical feature aggregation scheme that captures the natural hierarchical structure of panoramic views.
\item We introduce an adjustable hierarchical retrieval mechanism that selectively activates descriptors at different hierarchy levels, allowing flexible control over the accuracy–efficiency trade-off.
\item We validate the effectiveness and generality of our approach through extensive experiments across various datasets and configurations.
\end{itemize}

\section{Related Works}
\label{sec:related works}

\noindent\textbf{Visual place recognition.\hspace{0.15cm}} 
Visual Place Recognition (VPR) aims to identify the location of a query image by matching it against a reference image database. Recent VPR models~\cite{hausler2021patch,wang2022transvpr,zhu2023r2former,shen2023structvpr,izquierdo2024close,ali2024boq,lu2024towards,lu2024cricavpr} mainly focus on P2P matching between query and database images. 
However, this requires storing multiple perspective images for each location, which increases memory usage and hinders scalability in practical applications such as mobile or edge deployment. 
Equirectangular-to-equirectangular methods~\cite{fang2020cfvl,cheng2019panoramic,wang2018omnidirectional} alleviate this by using panoramic images, but they assume panoramic queries, introducing another limitation.

P2E VPR~\cite{orhan2021efficient,shi2023panovpr} provides a practical alternative by using perspective queries with panoramic databases.
However, existing methods typically decompose panoramas into perspective-view crops and employ sliding-window comparisons to address FoV mismatches. This approach yields limited efficiency and neglects the intrinsic structural relationships within panoramas.
To address these limitations, we propose HypeVPR, which learns hierarchical embeddings in hyperbolic space.
Our framework encodes global and local information within a single multi-level descriptor, enabling efficient hierarchical matching that drastically reduces window-based comparisons while preserving high accuracy.

\vspace{0.2cm}
\noindent\textbf{Hyperbolic manifolds.\hspace{0.15cm}} 
Hyperbolic manifolds have gained significant attention for effectively modeling hierarchical structures. Hyperbolic spaces are naturally suited to embed hierarchies (\eg tree graphs) with low distortion~\cite{sarkar2011low,sala2018representation}.
Initially popularized in NLP~\cite{nickel2017poincare,nickel2018learning}, hyperbolic embeddings have recently gained attention in vision tasks, including image retrieval~\cite{ermolov2022hyperbolic,kim2023hier}, segmentation~\cite{weng2021unsupervised,atigh2022hyperbolic}, and few-shot learning~\cite{gao2021curvature}. Khrulkov \textit{et al.}~\cite{HyperbolicImageEmbeddings} showed their benefits over Euclidean embeddings for hierarchical visual data, and Desai \textit{et al.}~\cite{desai2023hyperbolic} extended these ideas to multi-modal settings.

We build on these foundations by applying hyperbolic embeddings to panoramic image descriptors. By hierarchically aggregating multi-scale features in hyperbolic space, our method captures both local details and global context in a compact representation. Unlike prior work, we explicitly model the geometric structure of panoramic images through learned hierarchy, enabling descriptors that are both discriminative and generalizable. This structure improves robustness to FoV variation and enables fast, scalable retrieval for P2E VPR.

\section{Preliminaries}
\label{sec:Preliminaries}

\subsection{Hyperbolic geometry}
Formally, an $n$-dimensional hyperbolic space $\mathbb{H}^n$ is a simply connected Riemannian manifold with constant negative sectional curvature. While it cannot be isometrically embedded in Euclidean space, it shares certain structural properties with Euclidean spheres~\cite{krioukov2010hyperbolic, linial1998low}. Among the five well-studied isomorphic models of hyperbolic geometry, we adopt the Poincaré ball model for HypeVPR due to its extensive use in representation learning~\cite{HNN,HyperbolicImageEmbeddings,nickel2017poincare,ge2023hyperbolic,li2023euclidean}.

\vspace{0.2cm}
\noindent\textbf{Poincaré ball model.\hspace{0.15cm}} The Poincaré ball model $(\mathbb{D}^n, g^{\mathbb{D}})$ is defined by the manifold
$
\mathbb{D}^n = \left\{ \mathbf{x} \in \mathbb{R}^n : \|\mathbf{x}\| < 1 \right\}
$
equipped with the \textit{Riemannian metric} $g_{\mathbf{x}}^{\mathbb{D}} = {\lambda_{\mathbf{x}}^c}^2 g^E$, where 
\begin{equation}
   \lambda^c_{\mathbf{x}} := \frac{2}{1 - c\| \mathbf{x} \|^2} 
\end{equation} 
is the conformal factor and $g^E$ is the Euclidean metric tensor $g^E = I^n$. In this model, the \textit{geodesic distance} between two points is given by the following expression:
\begin{equation}
\label{eq:geodesic}
    d_{\mathbb{D}}(\mathbf{x}, \mathbf{y}) = \operatorname{arccosh} \left( 1 + 2 \frac{\|\mathbf{x} - \mathbf{y}\|^2}{(1 - \|\mathbf{x}\|^2)(1 - \|\mathbf{y}\|^2)} \right).
\end{equation}
In the Poincaré ball model, the norm of a feature vector reflects its level of semantic hierarchy; 
features with larger norms correspond to more fine-grained and distinctive representations, 
whereas those closer to the origin represent more general and abstract concepts.

\subsection{Hyperbolic Operations}
We use the generalized formula for operation on the Poincaré ball with an additional curvature parameter c, which modifies the ball's curvature, following~\cite{HNN,HyperbolicImageEmbeddings}.

\vspace{0.2cm}
\noindent\textbf{Distance.\hspace{0.15cm}}The distance function with curvature c in Poincaré ball is defined as 
\begin{equation}
\label{eq:distance}
    d_c(\mathbf{x}, \mathbf{y}) := \frac{2}{\sqrt{c}} \operatorname{arctanh}(\sqrt{c} \|\mathbf{x} \oplus_c \mathbf{y}\|),
\end{equation}
where $\oplus_c$ is Möbius addition with curvature c.

  



\vspace{0.2cm}
\noindent\textbf{Exponential and logarithmic maps.\hspace{0.15cm}}
To operate in hyperbolic space, a bijective mapping between \(\mathbb{R}^n\) and \(\mathbb{D}_c^n\) is required to convert Euclidean vectors to hyperbolic coordinates and back. 
This bijection is provided by the exponential and logarithmic maps, which map Euclidean vectors to hyperbolic space and serve as their inverse, respectively.

The \textit{exponential} map $\exp_{\mathbf{x}}^c$ is a function from tangent space $T_{\mathbf{x}} \mathbb{D}_c^n \cong \mathbb{R}^n$ to $\mathbb{D}_c^n$, defined as
\begin{equation}
\label{eq:expmap}
    \exp_{\mathbf{x}}^c(\mathbf{v}) := \mathbf{x} \oplus_c \left( \tanh \left( \frac{\sqrt{c} \lambda_{\mathbf{x}}^c \|\mathbf{v}\|}{2} \right) \frac{\mathbf{v}}{\sqrt{c} \|\mathbf{v}\|} \right).
\end{equation}

The inverse \textit{logarithmic} map is defined as
\begin{equation}
\label{eq:logmap}
    \log_{\mathbf{x}}^c(\mathbf{y}) := \frac{2}{\sqrt{c} \lambda_{\mathbf{x}}^c} \operatorname{arctanh}(\sqrt{c} \| -\mathbf{x} \oplus_c \mathbf{y} \|) \frac{-\mathbf{x} \oplus_c \mathbf{y}}{\| -\mathbf{x} \oplus_c \mathbf{y} \|}.
\end{equation}

In practice, the maps $\exp_0^c$ and $\log_0^c$ are utilized to transition between Euclidean and Poincaré ball representations of a vector.

\begin{figure*}
  \centering
    \includegraphics[width=1.0\linewidth]{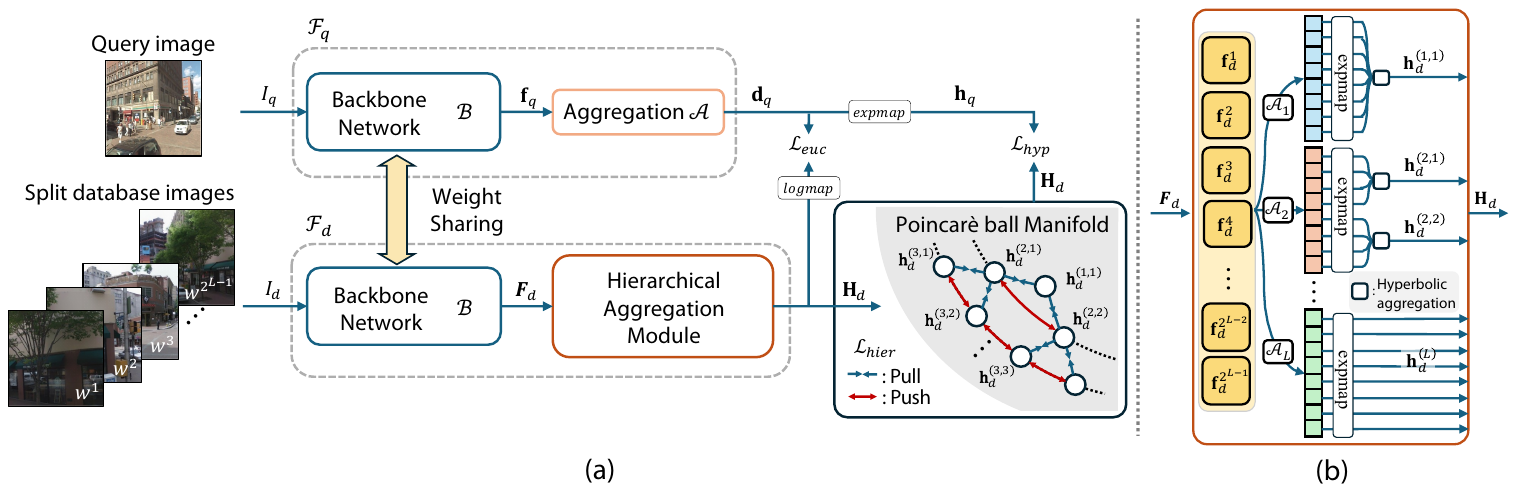}
  \caption{(a) Structure of our network and our training scheme. (b) Illustration of the Hierarchical Aggregation Module (HAM).}
  \label{fig:Model}
\end{figure*}

\section{Method}
\label{sec:method}

In this section, we propose the Hyperbolic P2E Visual Place Network (HypeVPR). The key aspect of HypeVPR is its ability to output visual descriptors from equirectangular images that can be directly compared with those from perspective view (PV) images.

\subsection{Overview}

Our framework follows the standard VPR pipeline, where a network \( \mathcal{F} \) extracts a fixed-size descriptor \( \mathbf{d} = \mathcal{F}(I) \in \mathbb{R}^{C_d} \) from both the query and database images. 
For a perspective query \( I_q \in \mathbb{R}^{H \times W \times C} \), the goal is to retrieve the most relevant geotagged database image \( I_{d} \in \mathcal{D} \subset \mathbb{R}^{H \times W' \times C} \), with \( W' > W \). 
Since a perspective query captures only a limited FoV while a panorama spans the full horizontal field, bridging this FoV gap becomes essential.
HypeVPR addresses this by hierarchically modeling the equirectangular database images and embedding their features in hyperbolic space, enabling hierarchical relationships to be naturally represented through exponential distance scaling.

\subsection{Hierarchical modeling of equirectangular image}
\label{ssec:modeling}

To hierarchically model an equirectangular image in hyperbolic space, we define an \(L\)-level structure by progressively halving its horizontal field of view. The top level is \( I_d^{(1)} = I_d \), and each level \( \ell \) is given by
\begin{equation}
    I_{d}^{(\ell)} \in \mathbb{R}^{H \times \frac{W'}{2^{\ell-1}} \times C}.
    \label{eq:window}
\end{equation}

In practice, the query \(I_q\) is resized to \(W \times W\), and the panorama \(I_d\) to \(W \times (8W)\), so that the lowest-level segments \(I_d^{(L)}\) match the query resolution and can be encoded by the same backbone. 
For \(L > 4\), the lowest-level windows overlap horizontally; we address this using a sub-tree partitioning strategy (see the supplementary material).

The resulting descriptors are aggregated from lower to higher levels form a hierarchical tree that captures both local details and global context for robust P2E matching.

\subsection{Network for Query Descriptors}

The network \( \mathcal{F}_q \) in the first path of Fig.~\ref{fig:Model} generates a Euclidean descriptor from the query image \( I_q \), consisting of a backbone \( \mathcal{B} \) and an aggregator \( \mathcal{A}_q \). The backbone extracts features \( \mathbf{f}_q = \mathcal{B}(I_q) \), which are then transformed into the final descriptor by the aggregator. The overall process is:
\begin{equation}
    \mathbf{d}_q = \mathcal{F}_q(\textit{I}_q) = \mathcal{A}_q(\mathcal{B}(\textit{I}_q)).
\end{equation}

For aggregation, we employ GeM pooling~\cite{gem}, which adaptively captures different types of information depending on its learnable parameter, followed by a linear projection layer:
\begin{equation}
\label{eq:aggregator}
\mathbf{d}_q = \mathcal{A}_q(\mathbf{f}_q) = \text{Linear}_q(\text{GeM}_q(\mathbf{f}_q)).
\end{equation}

To embed the Euclidean space descriptor \( \mathbf{d}_q \) into the hyperbolic descriptor embedding \( \mathbf{h}_q \), we use Equation~\eqref{eq:expmap} as follows:
\begin{equation}
\label{eq:expmap0_query}
    \mathbf{h}_q = \exp_{0}^c(\mathbf{d}_q) = \tanh \left( \sqrt{c} \|\mathbf{d}_q\| \right) \frac{\mathbf{d}_q}{\sqrt{c} \|\mathbf{d}_q\|}.
\end{equation}
This hyperbolic descriptor \( \mathbf{h}_q \) is used for matching with the database descriptors.

  


\subsection{Hierarchical aggregation network for database descriptors}

The network \( \mathcal{F}_d \) in the second path of Fig.~\ref{fig:Model} generates a descriptor \( \mathbf{H}_d \) from a database image \( I_d \). A key challenge is that \( \mathbf{H}_d \) must match the size of the query descriptor \( \mathbf{h}_q \) while encoding significantly more information. To address this, we design a hierarchical aggregation module (HAM) for the network \( \mathcal{F}_d \), which will be detailed in Sec.~\ref{sec:HAM}.

Since $\mathcal{F}_q$ and $\mathcal{F}_d$ share the same backbone $\mathcal{B}$, the output from the network $\mathcal{F}_d$ is represented by 
\begin{equation}
    \textbf{H}_d = \mathcal{F}_d(I_d) =\text{HAM}(\mathcal{B}(I_d))=\text{HAM}(\mathbf{F}_d).
\end{equation}
As we mentioned in Sec.~\ref{ssec:modeling}, first we divide database image $I_d$ into $2^{L-1}$ windows for the $L$-th level.  $I_d = \{ w^j : j = 1, \dots, 2^{L-1} \}$ and extract feature from each window by
\begin{equation}
\mathbf{F}_d = \{ \mathbf{f}_{d}^j = \mathcal{B}(w^j) : w^j \in I_d \}.
\end{equation}

To generate the hierarchical descriptor set \( \mathbf{H}_d \), we aggregate the window-based features \( \mathbf{f}_d^j \) at each level using HAM.

\subsection{Hierarchical aggregation module (HAM)}
\label{sec:HAM}
HAM aggregates Euclidean features into a cascaded hierarchy of hyperbolic descriptors across \(L\) levels (Fig.~\ref{fig:Model}(b)). 

Given the database features 
\(
\mathbf{F}_d = \{\mathbf{f}_d^1, \mathbf{f}_d^2, \dots, \mathbf{f}_d^{2^{L-1}}\},
\)
a level-wise spatial aggregator \( \mathcal{A}_\ell \) generates Euclidean descriptors
\begin{equation}
    \mathbf{d}_d^{(\ell), j} = \mathcal{A}_\ell(\mathbf{f}_d^j),
\end{equation}
where each \( \mathcal{A}_\ell \) follows the structure in Eq.~\eqref{eq:aggregator} but uses independent parameters.  
These descriptors are then projected into hyperbolic space via the exponential map:
\begin{equation}
    \mathbf{h}_d^{(\ell), j} = \exp_0^c(\mathbf{d}_d^{(\ell), j}).
\end{equation}

Next, descriptors at level \(\ell\) are partitioned into  
\(k \in \{1, \dots, 2^{\ell-1}\}\) non-overlapping groups,  which share the same shape as \(I_d^{(\ell)}\), each defined by
\begin{equation}
\mathcal{I}_d^{(\ell, k)} =
\left\{j \mid (k-1) \cdot 2^{L-\ell} + 1 \le j \le k \cdot 2^{L-\ell} \right\}.
\end{equation}
Each group is then aggregated into a single hyperbolic descriptor using the hyperbolic averaging operator:
\begin{equation}
    \mathbf{H}_d^{(\ell)} 
    = \left\{ 
        \mathbf{h}_d^{(\ell, k)} 
        = 
        \mathcal{A}_{hyp}
        \left(
            \{h_d^{(\ell), j} : j \in \mathcal{I}_d^{(\ell, k)}\}
        \right)
    \right\}.
\end{equation}
The top-level descriptor \( \mathbf{h}_d^{(1,1)} \) serves as the final representation of the database image and is directly compared with the query descriptor.

Since descriptors within a level have different norms reflecting their semantic hierarchy, aggregation must respect hyperbolic geometry. 
We therefore use the Einstein midpoint on the Klein model~\cite{HyperbolicImageEmbeddings}:
\begin{equation}
\label{eq:hypave}
    \mathcal{A}_{hyp}(h_1, \dots, h_n)
    = 
    \frac{\sum_{j=1}^{n} \gamma_j h_j}{\sum_{j=1}^{n} \gamma_j},
\end{equation}
where the Lorentz factors are
\begin{equation}
\label{eq:lorentz}
    \gamma_j = \frac{1}{\sqrt{1 - c\|h_j\|^2}}.
\end{equation}
This norm-aware weighting preserves the geometry of the Poincaré ball and enables effective hyperbolic feature aggregation. Additional details on the Klein model are provided in the supplementary material.

\subsection{Adjustable hierarchical retrieval}
\label{ssec:rerank}
Our hierarchical structure enables flexible multi-level matching, where coarse-to-fine descriptors can be selectively utilized to balance accuracy and efficiency. Instead of relying solely on the top-level descriptor, lower-level descriptors are used to refine the initial retrieval results.

We first compute hyperbolic distances between the query descriptor and the first-level database descriptors $\mathbf{h}_d^{(1,1)}$ by Eq.~\eqref{eq:distance} and retrieve the top-\(K'\) candidates.

After that, descriptors from selected levels \(\mathbb{L} \subset \{2, \dots, L\}\) are used to rescore only these \(K'\) candidates. At each level \(\ell\), we compute the minimum hyperbolic distance across all sub-descriptors:
\begin{equation}
    d_\ell = \min_{k} d_c(\mathbf{h}_q, \mathbf{h}_d^{(\ell,k)}).
\end{equation}
These level-wise distances are then normalized using Z-score normalization to ensure comparability:
\begin{equation}
\label{eq:zscore}
    \hat{s}_\ell = -\frac{d_\ell - \mu_\ell}{\sigma_\ell + \epsilon},
\end{equation}
where \(\mu_\ell\) and \(\sigma_\ell\) are computed over the \(K'\) candidates at level \(\ell\). The negative sign assigns higher scores to smaller distances.

 The final reranking score is obtained as a weighted sum over the selected levels:
\begin{equation}
\label{eq:rerank_zscore}
    s = \sum_{\ell \in \{1\} \cup \mathbb{L}} w_\ell \, \hat{s}_\ell,
\end{equation}
and the final top-\(K\) results are produced by sorting these scores in descending order.

By controlling the selected level set \(\mathbb{L}\), our system flexibly balances accuracy and efficiency without additional training. In practice, multi-level score fusion improves retrieval performance over the lowest-level sliding-window baseline while running significantly faster.





\subsection{Training objectives}
\label{ssec:loss}
We train the descriptors using three loss functions based on the triplet loss~\cite{balntas2016learning}.

\vspace{0.2cm}
\noindent\textbf{Hierarchical triplet loss.\hspace{0.15cm}}
To learn the hierarchical organization of \( \mathbf{H}_d \), we draw inspiration from the hierarchical contrastive loss~\cite{kwon2024improving} and adapt it to better reflect the scene structure. 
Rather than relying solely on level-wise similarity, we define positive and negative relations based on the spatial hierarchy: 
descriptors with overlapping FoVs across adjacent levels are treated as positives, while those from distinct regions within the same level serve as negatives, as illustrated in Fig.~\ref{fig:Model}. 
This encourages smooth transitions from local to global semantics and enforces geometric consistency across levels.
The loss is computed using the hyperbolic distance \( d_c \) in Equation~\eqref{eq:distance}:
\begin{align}
\mathcal{L}_{\text{hier}} = \sum\sum_{\mathbf{H}_d} \max \Big\{ 
& d_c\left(\mathbf{h}_{d}^{(\ell-1,k)}, \mathbf{h}_{d}^{(\ell,2k-b)}\right) \notag \\
& - d_c\left(\mathbf{h}_{d}^{(\ell,2k-b)}, \mathbf{h}_{d}^{(\ell,n)}\right) + m, 0 \Big\}.
\end{align}
Here, $\ell$ denotes the hierarchy level ranging from 2 to $L$, \( b \in \{0,1\} \) indexes child nodes, and \( n \) iterates over all other descriptors at level \( \ell \) as negatives. \( m \) represents the margin parameter.

\begin{table*}[t]
\caption{Performance comparison on Pitts250K-P2E and YQ360 datasets using P2E baselines. The first block includes methods that encode an entire panoramic image into a single descriptor, while the second block contains methods that represent a panoramic image using multiple descriptors. ‘*’ indicates models trained on an extra-large dataset. The subscript next to PanoVPR denotes the number of sliding-window crops used to represent each panoramic database image. Our models are highlighted in gray.}
\centering
\small
\setlength{\tabcolsep}{4pt}
\resizebox{0.98\textwidth}{!}{
\begin{tabular}{l|c|cccc|cccc|c}
\toprule
\multirow{2}{*}{Method} & \multirow{2}{*}{Backbone}
& \multicolumn{4}{c|}{Pitts250K-P2E} 
& \multicolumn{4}{c|}{YQ360} 
& \multirow{2}{*}{\#Params. (M)} \\
\cmidrule(lr){3-6} \cmidrule(lr){7-10}
& 
& Time/q (ms) & R@1 & R@5 & R@10 
& Time/q (ms) & R@1 & R@5 & R@10 
&  \\
\midrule

SwinT~\cite{liu2021swin} 
& Swin-T
& \textbf{1.2} & 10.1 & 26.3 & 36.0 
& \textbf{0.7} & 27.4 & 63.5 & 72.3 
& 28.29 \\

ConvNeXtS~\cite{liu2022convnet} 
& ConvNeXt-S
& \textbf{1.2} & 14.2 & 28.8 & 39.6 
& \textbf{0.7} & 37.0 & 72.5 & 83.9 
& 50.22 \\

NetVLAD~\cite{arandjelovic2016netvlad} 
& ResNet-50
& 32.8 & 4.0 & 12.4 & 20.0 
& 19.1 & 35.2 & 66.8 & 80.0 
& 25.30 \\

Berton \etal (ViT-B)~\cite{berton2022deep} 
& ViT-B
& \textbf{1.2} & 8.0 & 23.0 & 33.0 
& \textbf{0.7} & 40.4 & 74.8 & 88.4 
& 86.86 \\

\rowcolor[HTML]{F2F2F2}
HypeVPR-O$^{*}$
& ResNet-50
& 4.0 & 66.5 & 82.1 & 86.3 
& 1.9 & 53.6 & 81.2 & 88.8 
& 27.70 \\
\midrule

PanoVPR\textsubscript{$\times$8}~\cite{shi2023panovpr} 
& Swin-T
& 17.0 & 22.0 & 42.2 & 51.8 
& 4.9 & 30.8 & 69.6 & 81.6 
& 28.29 \\

\rowcolor[HTML]{F2F2F2}
HypeVPR-B 
& Swin-T
& \underline{3.6} & 29.4 & 51.4 & 60.6 
& \underline{4.0} & 38.0 & 74.4 & 88.0 
& 28.29 \\

PanoVPR\textsubscript{$\times$8}~\cite{shi2023panovpr} 
& ConvNeXt-S
& 17.0 & 30.9 & 53.9 & 64.3 
& 4.9 & 39.6 & 76.8 & 87.6 
& 50.22 \\

\rowcolor[HTML]{F2F2F2}
HypeVPR-B
& ConvNeXt-S
& \underline{3.6} & 34.3 & 59.0 & 71.1 
& \underline{4.0} & 43.8 & 78.8 & 89.2 
& 50.22 \\

PanoVPR\textsubscript{$\times$16}~\cite{shi2023panovpr} 
& Swin-T
& 48.6 & 33.6 & 56.7 & 66.4 
& 11.0 & 43.2 & 82.4 & 90.8 
& 28.29 \\

\rowcolor[HTML]{F2F2F2}
HypeVPR-L
& Swin-T
& 14.0 & 32.5 & 57.2 & 67.6 
& 6.7 & 45.6 & 84.0 & 91.2 
& 28.29 \\

PanoVPR\textsubscript{$\times$16}~\cite{shi2023panovpr} 
& ConvNeXt-S
& 48.6 & 40.3 & 63.0 & 72.1 
& 11.0 & 46.0 & 83.2 & 92.4 
& 50.22 \\

\rowcolor[HTML]{F2F2F2}
HypeVPR-L 
& ConvNeXt-S
& 14.0 & 43.4 & 64.3 & 73.4 
& 6.7 & \underline{52.4} & \underline{85.2} & \underline{94.8} 
& 50.22 \\

Orhan \etal~\cite{orhan2021efficient}$^{*}$ 
& ResNet-101
& 1555.2 & \underline{47.0} & \underline{66.4} & \underline{73.6} 
& 981.4 & 47.6 & 79.2 & 88.4 
& 136.62 \\

\rowcolor[HTML]{F2F2F2}
HypeVPR-B$^{*}$  
& ResNet-50
& 29.6 & \textbf{79.6} & \textbf{88.6} & \textbf{90.6} 
& 14.2 & \textbf{63.6} & \textbf{88.8} & \textbf{96.4} 
& 27.70 \\
\bottomrule
\end{tabular}
}

\label{tab:comparison}
\end{table*}

\vspace{0.2cm}
\noindent\textbf{Hyperbolic triplet loss.\hspace{0.15cm}}
We use $\mathcal{L}_{\text{hyp}}$ for matching the representative descriptor $\mathbf{h}_{d}^{(1,1)}$ of $\mathbf{H}_d$ and the query descriptor $\mathbf{h}_q$. $\mathcal{L}_{\text{hyp}}$ is defined using distance metric $d_c$ as follows:
\begin{align}
\label{eq:hyp_triplet}
\mathcal{L}_{\text{hyp}} = \sum \max\Big\{ 
    & d_c\left(\mathbf{h}_q, \mathbf{h}_{d,\mathbb{P}}^{(1,1)}\right) \notag \\
    & - d_c\left(\mathbf{h}_q, \mathbf{h}_{d,\mathbb{N}}^{(1,1)}\right) + m, 0 \Big\},
\end{align}
where \( \mathbb{P}' \) and \( \mathbb{N}' \) denote the positive and negative samples selected from the sets \( \mathbb{P} \) and \( \mathbb{N} \), respectively, using the mining method described in~\cite{shi2023panovpr}. The sets \( \mathbb{P} \) and \( \mathbb{N} \) are determined before training.

\vspace{0.2cm}
\noindent\textbf{Euclidean triplet loss.\hspace{0.15cm}}
To stabilize training and ensure that window-based features are properly learned, 
we additionally apply a Euclidean triplet loss \( \mathcal{L}_{\text{euc}} \) on the query descriptor and the lowest-level (\(L\)) database descriptor. 
Specifically, we map the hyperbolic descriptors \( \mathbf{h}_q \) and \( \mathbf{H}^{(L)}_d \) back to Euclidean space via the logarithmic map in Equation~\eqref{eq:logmap}, 
and compute \( \mathcal{L}_{\text{euc}} \) using the \( L_2 \) distance as follows:
\begin{align}
\label{eq:euc_triplet}
\mathcal{L}_{euc} = 
\sum \max \bigg\{ 
& \| \mathbf{d}_q - \mathbf{d}_{d,\mathbb{P}'}^{(L)} \|_2 \nonumber \\
& - \| \mathbf{d}_q - \mathbf{d}_{d,\mathbb{N}'}^{(L)} \|_2 + m, \; 0 
\bigg\},
\end{align}
where $\mathbb{P}$ and $\mathbb{N}$ denote the same positive and negative sets used in $\mathcal{L}_{\text{hyp}}$.

\vspace{0.2cm}
\noindent\textbf{Overall objectives.  }
The overall objectives are defined as:
\begin{equation}
    \mathcal{L}=\mathcal{L}_{hier}+\mathcal{L}_{hyp}+\mathcal{L}_{euc}.
\end{equation}

\section{Experiments}
\label{sec:experiments}
In this section, we conduct extensive experiments to demonstrate the effectiveness of our proposed HypeVPR for the P2E VPR task.

\subsection{Implementation Details}
\noindent\textbf{Training.\hspace{0.15cm}}
We train the model with a batch size of 2 using the RiemannianAdam optimizer~\cite{kochurov2020geoopt}, with a learning rate of 1e\text{-}5 and a triplet loss margin \(m = 0.1\). Training runs for up to 60 epochs with early stopping after 10 epochs without validation improvement.

Following standard VPR practice~\cite{arandjelovic2016netvlad}, we mine 1 hard positive and 10 hard negatives per query using KNN over GPS coordinates, reducing the positive radius to 10\,m. Partial mining~\cite{berton2022deep} is applied to sample a subset of candidates. All query images are resized to \(224 \times 224\), and database panoramas are resized so that the lowest level corresponds to \(W' = 224 \times 8\).

We use the Poincaré ball model with fixed curvature \(c = 1\). All experiments are conducted on a single NVIDIA A5000 GPU.

\vspace{0.2cm}
\noindent\textbf{Model setting.\hspace{0.15cm}}
We conduct experiments under two configurations to ensure a fair comparison between P2E and P2P settings. 
For P2E, perspective queries are resized to \(W = H = 224\), and database panoramas are resized to \(W' = 224 \times 8\), following the training setup. 
For P2P, queries are resized to \(W = H = 512\), and database images are set to \(W' = 512 \times 8\), consistent with standard P2P-based VPR. 
We use a descriptor dimension of 768 for P2E and 2048 for P2P to match their respective conventions.

Because panorama width satisfies $W' = 8W$, the lowest-level sub-windows become narrower as the hierarchy depth $L$ increases. 
For $L > 4$, the lowest level windows overlap horizontally, which disrupts the positive–negative relationships required by the triplet loss. 
To avoid this issue, we apply a sub-tree partitioning strategy (see supplementary material). We set $L = 5$ unless otherwise stated.

For adjustable retrieval, we adopt the weighting scheme
that yields the highest validation accuracy. To explicitly ablate the effect of different hierarchy levels and the number of windows, we construct several variants. 
We denote the model using only \(\mathbf{h}_d^{(1)}\) as HypeVPR-O (one), 
using \(\mathbf{h}_d^{(1)}\) and \(\mathbf{h}_d^{(4)}\) as HypeVPR-B (base),
using \(\mathbf{h}_d^{(1)}\) and \(\mathbf{h}_d^{(5)}\) as HypeVPR-L (large),
and using only \(\mathbf{h}_d^{(5)}\) as HypeVPR-SW (sliding window).

Additional details on training, model settings, datasets, and evaluation metrics are provided in the supplementary material.

\subsection{Comparison with P2E baselines}
Tab.~\ref{tab:comparison} compares our method with state-of-the-art approaches for P2E VPR on the Pitts250K-P2E and YQ360~\cite{shi2023panovpr} datasets.
All methods are evaluated under the same settings as PanoVPR to ensure a fair comparison with P2E baselines.

\begin{figure}[t]
  \centering
   \includegraphics[width=1\linewidth]{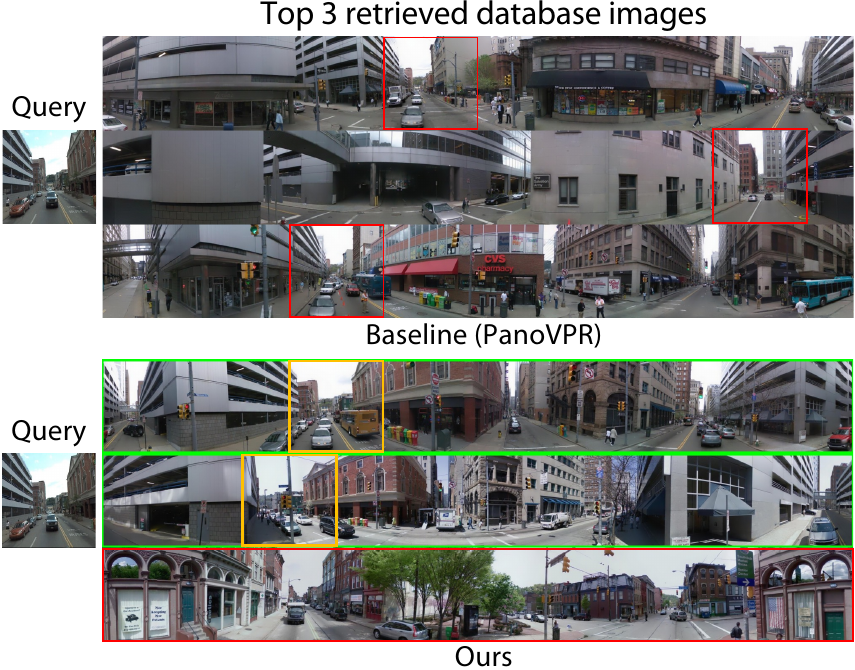}
   \caption{Qualitative results. Correct predictions are outlined in green, while incorrect predictions are outlined in red. In our method, the window with the highest score within the positive pair is highlighted in yellow. Best viewed when zoomed in.}
   \label{fig:qualitative}
\end{figure}

Our method achieves superior performance across all benchmarks and configurations on both datasets.
This advantage is evident not only over methods that encode a panoramic image into a single descriptor—such as NetVLAD~\cite{arandjelovic2016netvlad} and Berton et al.~\cite{berton2022deep}—but also over sliding window-based approaches including PanoVPR~\cite{shi2023panovpr} and Orhan \etal~\cite{orhan2021efficient}.
On Pitts250K-P2E, our method consistently achieves the highest recall across all configurations, offering the best balance between accuracy and efficiency.
On YQ360, although the speed advantage narrows due to the smaller database size, our model still outperforms others across different backbones and window settings while maintaining sub-millisecond query time.
Qualitative examples in Fig.~\ref{fig:qualitative} further demonstrate its robustness over existing P2E baselines, with more results in the supplementary material.

Note that the reported retrieval time measures only the matching stage per query on CPU—excluding feature extraction—and thus depends solely on the descriptor dimension, the number of descriptors compared, and any additional computational overhead.

\subsection{Comparison with P2P baselines and trade-off control}
To ensure a fair comparison with P2P baselines trained on large-scale datasets, we adopt EigenPlace~\cite{berton2022rethinking} as our backbone and fine-tune it with HAM on Pitts250k-P2E to enable hierarchical embedding.
By maintaining identical window sizes and backbone architectures for both query and database networks, our model can leverage P2P VPR's pretrained backbones and well-established training strategies as effective priors.
We further evaluate state-of-the-art P2P methods by extending them to panoramic images divided into 16 overlapping crops following~\cite{shi2023panovpr}.

As shown in Tab.~\ref{tab:model_comparison}, our method achieves accuracy comparable to prior models while offering substantial gains in retrieval speed and memory efficiency. 
HypeVPR-B achieves an R@1 of 79.6—on par with EigenPlace~\cite{berton2023eigenplaces}—while delivering over 5$\times$ faster retrieval and nearly 2$\times$ lower storage usage.
HypeVPR-L attains an R@1 of 81.2, surpassing EigenPlace while still maintaining over 2$\times$ faster retrieval.
Notably, combining the scores of $\small\mathbf{h}_d^{(1)}$ and $\small\mathbf{h}_d^{(5)}$ (HypeVPR-L) yields higher performance than using only $\small\mathbf{h}_d^{(5)}$ (HypeVPR-SW) in the exhaustive sliding-window setting. 
Even when using only $\small\mathbf{h}_d^{(1)}$ (HypeVPR-O), our model achieves over 61$\times$ faster retrieval and requires 66$\times$ less storage than SALAD~\cite{izquierdo2024optimal}.

We further evaluate our method on the SF-XL dataset~\cite{berton2022rethinking}, as shown in Tab.~\ref{tab:sf_xl}. 
To this end, we construct a panoramic test database by replacing each test PV image with its corresponding panoramic version, yielding 0.23M panoramas that collectively cover the 2.8M PV images in the SF-XL test set. 
This conversion highlights the advantage of maintaining the database in panoramic form, which substantially reduces redundancy while preserving full scene coverage. 
Despite being trained only on the smaller Pitts250k-P2E dataset, our model maintains robust performance on the large-scale SF-XL domain, outperforming most existing methods.
Although slightly below SALAD~\cite{izquierdo2024optimal} in recall, HypeVPR-L is 11$\times$ faster and requires only one-third of its storage.

\begin{table} 
\caption{Comparison of state-of-the-art P2P VPR methods on Pitts250k-P2E dataset. * indicates models fine-tuned on Pitts250k-P2E. } 
\centering 
\setlength{\tabcolsep}{2pt} 
\small 
\begin{tabular}{lcccc} 
\toprule 
Model & time/q (ms) & Storage (MB) & Desc. & R@1 \\
\midrule 
EigenPlace~\cite{berton2023eigenplaces} & 90.5 & 262.4 & 2048 & 78.3 \\ 
EigenPlace*~\cite{berton2023eigenplaces} & 90.5 & 262.4 & 2048 & 80.9 \\ CosPlace~\cite{berton2022rethinking} & 90.5 & 262.4 & 2048 & 73.1 \\ ConvAP~\cite{ali2022gsv} & 90.5 & 262.4 & 2048 & 72.8 \\ 
SALAD~\cite{izquierdo2024optimal} & 371.4 & 1082.4 & 8448 & \textbf{86.8} \\ 
\midrule 
HypeVPR-O* & \textbf{6.1} & \textbf{16.4} & 2048 & 66.5 \\ 
HypeVPR-SW* & 90.5 & 262.4 & 2048 & 80.6 \\ 
HypeVPR-B* & \underline{17.9} & \underline{147.6} & 2048 & 79.6 \\ 
HypeVPR-L* & 41.8 & 278.8 & 2048 & \underline{81.2} \\ 
\bottomrule 
\end{tabular} 
\label{tab:model_comparison} 
\end{table}

\begin{table}[t] 
\caption{Performance comparison of state-of-the-art P2P VPR methods on the SF-XL test set using query v1.} 
\centering 
\small 
\setlength{\tabcolsep}{3pt} 
\begin{tabular}{lcccc} 
\toprule Model & time/q (s) & Storage (GB) & Desc. & R@1 \\ 
\midrule 
CosPlace~\cite{berton2022rethinking} & 7.92 & 21.4 & 2048 & 76.4 \\ 
CricaVPR~\cite{lu2024cricavpr} & 15.85 & 42.8 & 4096 & 80.6 \\ 
EigenPlace~\cite{berton2023eigenplaces} & 7.92 & 21.4 & 2048 & 84.1 \\ 
BoQ~\cite{ali2024boq} & 63.4 & 171.3 & 16384 & 83.7 \\ 
SALAD-slim~\cite{izquierdo2024optimal} & 8.19 & 22.1 & 2112 & 86.5 \\
SALAD~\cite{izquierdo2024optimal} & 32.69 & 88.3 & 8448 & \textbf{88.6} \\ 
\midrule 
HypeVPR-B & \textbf{1.79} & \underline{16.0} & 2048 & 80.5 \\ 
HypeVPR-L & \underline{2.92} & 30.3 & 2048 & \underline{85.2} \\ 
\bottomrule 
\end{tabular} 
\label{tab:sf_xl} 
\end{table}


\subsection{Ablation studies}
We conducted an ablation study to demonstrate the effectiveness of each component of our framework. All experiments were performed on the Pitts250k-P2E dataset~\cite{shi2023panovpr}.

\noindent\textbf{Effect of hyperbolic manifold.\hspace{0.15cm}}
We evaluate the representational power of hyperbolic space by comparing two global descriptors: one from GeM pooling~\cite{gem} in Euclidean space, and another aggregated in hyperbolic space via HAM. 
This setting forms a simple two-level hierarchy—query-sized and database-sized windows.  
As shown in Tab.~\ref{tab:space}, even with this simple hierarchy, hyperbolic features significantly outperform Euclidean ones, supporting our design choice.

\vspace{0.2cm}
\noindent\textbf{Effect of each loss.\hspace{0.15cm}}
We evaluate the contributions of $\mathcal{L}_{\text{hier}}$, $\mathcal{L}_{\text{hyp}}$, and $\mathcal{L}_{\text{euc}}$ using the HypeVPR-O configuration under the same settings as Tab.~\ref{tab:model_comparison}. Tab.~\ref{tab:losses} reports the performance when each loss is removed. Removing $\mathcal{L}_{\text{euc}}$ severely degrades window-level feature learning, hindering hierarchical structure formation. Without $\mathcal{L}_{\text{hyp}}$, the model still performs reasonably well, indicating strong supervision provided by the hierarchy itself. Excluding $\mathcal{L}_{\text{hier}}$ lowers performance as the model must rely solely on descriptor matching without hierarchical guidance. Overall, the three losses are complementary, with the full model achieving the best performance.


\subsection{Feature visualization}
\begin{figure}[t]
  \centering
    \includegraphics[width=0.77\linewidth]{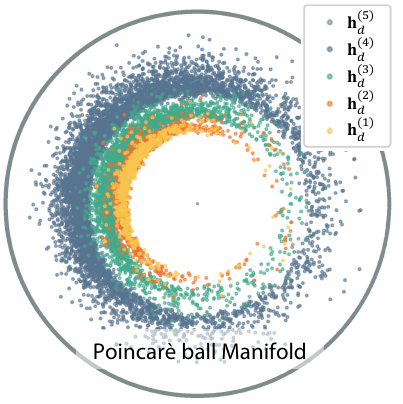}
  \caption{Visualization of 1,000 hierarchical descriptors of Pitts-250k P2E test set on the Poincaré ball manifold.
}
  \label{fig:feat_viz}
\end{figure}

To verify the hierarchical property of our hyperbolic embedding, 
we visualize the 1,000 Pitts250k-P2E~\cite{shi2023panovpr} test set descriptors on the Poincaré ball in terms of their norm and angular components (Fig.~\ref{fig:feat_viz}). 
Higher-level descriptors ($\mathbf{h}_d^{(1)}$) concentrate near the origin, indicating more abstract semantics, 
whereas lower-level descriptors appear closer to the boundary, capturing finer scene details. 
This distribution shows that our model organizes features according to their semantic hierarchy in hyperbolic space.


\begin{table}
\caption{Performance comparison of a 2-level hierarchy between Euclidean and Hyperbolic feature spaces.}
\centering
\setlength{\tabcolsep}{4pt}
\small
\begin{tabular}{l|c|c|c|c}
\toprule
Feature space & R@1 & R@5 & R@10 & R@20 \\
\midrule
Euclidean & 9.2 & 20.9 & 28.6 & 36.9 \\
Poincaré ball & \textbf{14.9} & \textbf{30.8} & \textbf{41.9} & \textbf{50.8} \\
\bottomrule
\end{tabular}

\label{tab:space}
\end{table}

\begin{table}
\caption{Effect of each loss. }
\centering
\setlength{\tabcolsep}{4pt}
\small
\begin{tabular}{l|c|c|c|c}
\toprule
Method & R@1 & R@5 & R@10 & R@20 \\
\midrule
w/o $\mathcal{L}_{\text{euc}}$ & 32.0 & 54.1 & 63.7 & 71.6 \\
w/o $\mathcal{L}_{\text{hyp}}$ & 64.3 & 81.1 & 86.0 & 88.8 \\
w/o $\mathcal{L}_{\text{hier}}$ & 50.8 & 75.6 & 81.6 & 85.7 \\
\midrule
Full model & \textbf{66.5} & \textbf{82.1} & \textbf{86.3} & \textbf{89.3} \\
\bottomrule
\end{tabular}

\label{tab:losses}
\end{table}

\vspace{-0.2cm}

\section{Conclusions}
\label{sec:conclusion}

In this paper, we introduced HypeVPR, a hyperbolic framework that models panoramic images through a hierarchical feature organization for Perspective-to-Equirectangular (P2E) Visual Place Recognition. By constructing a multi-level hierarchy from panoramic views and embedding it in hyperbolic space, our approach effectively captures both global contextual relations and fine-grained local details. This hierarchical structure enables adjustable hierarchical retrieval, providing flexible control over the trade-off between accuracy and efficiency. Extensive experiments demonstrate that HypeVPR achieves competitive performance with favorable efficiency trade-offs compared to existing methods. While hyperbolic embeddings currently lack compatibility with standard kNN search methods (e.g., FAISS~\cite{douze2024faiss}), addressing this limitation presents an exciting avenue for future P2E VPR research.

\noindent\textbf{Acknowledgments.} This work was supported by Korea Evaluation Institute Of
Industrial Technology (KEIT) grant funded by the Korea government(MOTIE)
(No.20023455).
 
{
    \small
    \bibliographystyle{ieeenat_fullname}
    \bibliography{main}
}



\end{document}